\documentclass[10pt,journal,compsoc]{IEEEtran}

\ifCLASSOPTIONcompsoc
  \usepackage[nocompress]{cite}
\else
  \usepackage{cite}
\fi
\ifCLASSINFOpdf
   \usepackage[pdftex]{graphicx}
\else
\fi
\usepackage{amsthm,amsmath,amssymb}
\usepackage{mathrsfs}
\usepackage{booktabs}
\usepackage{multirow}
\usepackage{multicol}
\usepackage{xspace}
\usepackage{bm}
\usepackage[dvipsnames,svgnames,x11names,table]{xcolor}
\usepackage[table]{xcolor}
\usepackage{hyperref}
\usepackage{xstring}

\usepackage{graphicx}
\usepackage{subcaption}
\usepackage{tikz}
\usetikzlibrary{patterns}
\usepackage{pgfplots}
\pgfplotsset{compat=newest}
\usetikzlibrary{pgfplots.groupplots}

\makeatletter
\renewcommand\paragraph{\@startsection{paragraph}{4}{\z@}{1.5ex}{-1em}{\normalfont\normalsize\bfseries}}
\makeatother

\usepackage{roboto}
\usepackage{pifont}

\usetikzlibrary{patterns.meta}
\pgfdeclarepattern{
  name=hatch,
  parameters={\hatchsize,\hatchangle,\hatchlinewidth},
  bottom left={\pgfpoint{-.1pt}{-.1pt}},
  top right={\pgfpoint{\hatchsize+.1pt}{\hatchsize+.1pt}},
  tile size={\pgfpoint{\hatchsize}{\hatchsize}},
  tile transformation={\pgftransformrotate{\hatchangle}},
  code={
    \pgfsetlinewidth{\hatchlinewidth}
    \pgfpathmoveto{\pgfpoint{-.1pt}{-.1pt}}
    \pgfpathlineto{\pgfpoint{\hatchsize+.1pt}{\hatchsize+.1pt}}
    \pgfpathmoveto{\pgfpoint{-.1pt}{\hatchsize+.1pt}}
    \pgfpathlineto{\pgfpoint{\hatchsize+.1pt}{-.1pt}}
    \pgfusepath{stroke}
  }
}

\tikzset{
  hatch size/.store in=\hatchsize,
  hatch angle/.store in=\hatchangle,
  hatch line width/.store in=\hatchlinewidth,
  hatch size=5pt,
  hatch angle=0pt,
  hatch line width=.5pt,
}

\tikzset{every mark/.append style={solid}}
\pgfplotsset{
	grid=both, width=\columnwidth, try min ticks=5,
	every axis/.append style={font=\small},
	every axis plot/.append style={thick,mark=none,mark size=1.5,tension=0.18},
	legend cell align=left, legend style={fill opacity=0.8},
	nodes near coords math/.style={
		nodes near coords={\pgfmathprintnumber[assume math mode=true]{\pgfplotspointmeta}},
	},
	every non boxed x axis/.style={},
}

\usepackage[ruled,vlined,linesnumbered]{algorithm2e}
\newcommand{\alert}[1]{{\color{red}{#1}}}

\newcommand{\Th}[1]{\textsc{#1}}
\newcommand{\mr}[2]{\multirow{#1}{*}{#2}}
\newcommand{\mc}[2]{\multicolumn{#1}{c}{#2}}
\newcommand{\tb}[1]{\textbf{#1}}
\newcommand{\ch}{\checkmark}

\newcommand{\citeme}[1]{\alert{[X]}}
\newcommand{\refme}[1]{\alert{(X)}}

\newcommand{\wb}[1]{\overline{#1}}
\newcommand{\wt}[1]{\widetilde{#1}}
\newcommand{\wh}[1]{\widehat{#1}}

\newcommand{\cM}{\mathcal{M}}

\newcommand{\cU}{\mathcal{U}}

\newcommand{\cX}{\mathcal{X}}

\newcommand{\vH}{\mathbf{H}}

\newcommand{\vX}{\mathbf{X}}

\newcommand{\vZ}{\mathbf{Z}}

\newcommand{\vc}{\mathbf{c}}

\newcommand{\vx}{\mathbf{x}}

\newcommand{\vz}{\mathbf{z}}

\newcommand{\vtheta}{{\boldsymbol{\theta}}}

\makeatletter
\newcommand*\bdot{\mathpalette\bdot@{.7}}
\newcommand*\bdot@[2]{\mathbin{\vcenter{\hbox{\scalebox{#2}{$\m@th#1\bullet$}}}}}
\makeatother

\makeatletter
\DeclareRobustCommand\onedot{\futurelet\@let@token\@onedot}
\def\@onedot{\ifx\@let@token.\else.\null\fi\xspace}

\def\eg{\emph{e.g}\onedot} 
\def\ie{\emph{i.e}\onedot}

\def\etal{\emph{et al}\onedot}
\makeatother

\newcommand{\tens}[1]{\bm{\mathsf{#1}}}

\newcommand{\tX}{\tens{X}}

\newcommand{\CSA}{\operatorname{CSA}}
\newcommand{\GCA}{\operatorname{GCA}}
\newcommand{\SA}{\operatorname{SA}}
\newcommand{\FFW}{\operatorname{FFW}}
\newcommand{\Norm}{\operatorname{Norm}}

\newcommand{\splt}{\operatorname{split}}

\newcommand{\ours}{DeepMLF\xspace}

\definecolor{Gray0}{gray}{0.4}
\definecolor{Gray1}{gray}{0.75}
\definecolor{Gray2}{gray}{0.80}
\definecolor{Gray2}{gray}{0.82}
\definecolor{Gray3-soft}{RGB}{220, 255, 255}
\definecolor{Gray3-dark}{RGB}{185, 221, 225}
\definecolor{Gray3}{RGB}{205, 246, 250} 
\definecolor{Gray4}{gray}{0.95}
\definecolor{my_Green}{RGB}{0,140,0}

\definecolor{LightBlue1}{RGB}{173, 216, 230} 
\definecolor{LightBlue2}{RGB}{135, 206, 250} 
\definecolor{LightBlue3}{RGB}{176, 224, 230} 

\newcommand{\blight}[1]{\textcolor{blue!50!white}{\textbf{#1}}} 
\newcommand{\bmedium}[1]{\textcolor{blue!75!white}{\textbf{#1}}} 
\newcommand{\bdark}[1]{\textcolor{blue}{\textbf{#1}}} 

\definecolor{CB91_Blue}{HTML}{2CBDFE}
\definecolor{CB91_Green}{HTML}{47DBCD}
\definecolor{CB91_Pink}{HTML}{F3A0F2}
\definecolor{CB91_Purple}{HTML}{9D2EC5}
\definecolor{CB91_Violet}{HTML}{661D98}
\definecolor{CB91_Amber}{HTML}{F5B14C}

\definecolor{Point_Color}{HTML}{9b59b6}
\definecolor{Hull_Line}{HTML}{34495e}
\definecolor{Combination}{HTML}{e67e22}

\begin{document}



\title{
\ours: Multimodal language model with learnable tokens for deep fusion \\ in sentiment analysis
}


%
%

\author{Efthymios~Georgiou,~\IEEEmembership{Graduate Student Member,~IEEE,}
        Vassilis~Katsouros,~~\IEEEmembership{Member,~IEEE,}
        Yannis~Avrithis,~\IEEEmembership{Senior Member,~IEEE,}
        and~Alexandros~Potamianos,~\IEEEmembership{Fellow,~IEEE}%
\IEEEcompsocitemizethanks{
\IEEEcompsocthanksitem Efthymios Georgiou is with the School of ECE, National Technical University of Athens, Athens, Greece, and the Institiute for Speech and Language Processing, Athena Research Center, Athens, Greece \protect\\
E-mail: efthygeo@mail.ntua.gr
\IEEEcompsocthanksitem Vassilis Katsouros is with the Institiute for Speech and Language Processing, Athena Research Center, Athens, Greece \protect\\
E-mail: vsk@athenarc.gr
\IEEEcompsocthanksitem Yannis Avrithis is with the Institute of Advanced Research on Artificial Intelligence (IARAI), Vienna, Austria \protect\\
E-mail: yannis@avrithis.net
\IEEEcompsocthanksitem Alexandros Potamianos is with the School of ECE, National Technical University of Athens, Athens, Greece \protect\\
E-mail: potam@central.ntua.gr
}
\thanks{Manuscript received xxx; revised xxx. (Corresponding author: Efthymios Georgiou)}
}

%
%


\markboth{Journal of \LaTeX\ Class Files,~Vol.~14, No.~8, August~2015}%
{Shell \MakeLowercase{\textit{et al.}}: Bare Demo of IEEEtran.cls for Computer Society Journals}

\IEEEtitleabstractindextext{%
\begin{abstract}
While multimodal fusion has been extensively studied in Multimodal Sentiment Analysis (MSA), the role of fusion depth and multimodal capacity allocation remains underexplored. In this work, we position fusion depth, scalability, and dedicated multimodal capacity as primary factors for effective fusion. We introduce \ours, a novel multimodal language model (LM) with learnable tokens tailored toward deep fusion. \ours leverages an audiovisual encoder and a pretrained decoder LM augmented with multimodal information across its layers. We append learnable tokens to the LM that: 1) capture modality interactions in a controlled fashion and 2) preserve independent information flow for each modality. These fusion tokens gather linguistic information via causal self-attention in LM Blocks and integrate with audiovisual information through cross-attention MM Blocks. Serving as dedicated multimodal capacity, this design enables progressive fusion across multiple layers, providing depth in the fusion process. Our training recipe combines modality-specific losses and language modelling loss, with the decoder LM tasked to predict ground truth polarity. Across three MSA benchmarks with varying dataset characteristics, \ours achieves state-of-the-art performance. Our results confirm that deeper fusion leads to better performance, with optimal fusion depths (5-7) exceeding those of existing approaches. Additionally, our analysis on the number of fusion tokens reveals that small token sets ($\sim$20) achieve optimal performance. We examine the importance of representation learning order (fusion curriculum) through audiovisual encoder initialization experiments. Our ablation studies demonstrate the superiority of the proposed fusion design and gating while providing a holistic examination of \ours's scalability to LLMs, and the impact of each training objective and embedding regularization.
\end{abstract}


\begin{IEEEkeywords}
Multimodal Learning, Deep Fusion, Learnable Tokens, Multimodal LM, LLMs, Multimodal Sentiment Analysis (MSA)
\end{IEEEkeywords}
}


\maketitle

\IEEEdisplaynontitleabstractindextext

%
\IEEEpeerreviewmaketitle

\IEEEraisesectionheading{
\section{Introduction}
\label{sec:introduction}
}

\IEEEPARstart{H}{umans} perceive and combine information from different sources and senses to understand and interact with their surroundings. Multimodal signals and representations are also utilized by the human brain when learning concepts. We can therefore claim that multimodality spans the entire human cognitive process.
Multimodal Machine Learning (MML) investigates how to develop systems or agents that can process and integrate heterogeneous and interconnected types of data, such as visual, auditory, and textual information.
The aim of this field involves the design of systems that understand, reason, and learn from the world through multiple sensory modalities, \eg, verbal and non-verbal communication and scene understanding.

From recognizing emotions through speech and language to generating images from text, the fundamental operation is \emph{multimodal fusion}~\cite{maragos2008multimodal}. Technically, fusion is the problem of learning representations that capture both unimodal information and cross-modal interactions between elements of different modalities. Conceptually, more homogeneous modalities are easier to combine compared to more heterogeneous. Fusion techniques can be broadly categorized into early, late, hybrid and deep fusion methods. Early fusion combines data at earlier stages, late fusion at the final stages, and hybrid fusion combines these schemes. Deep fusion typically involves multiple fusion stages within the architecture.

Recent works in the MML field employ deep fusion schemes to leverage the benefits of multimodality.
From the self-supervised approaches of ViLBERT~\cite{Lu2019ViLBERTPT} and UNITER~\cite{chen2020uniter} to multimodal large language model (LLM) based approaches~\cite{tsimpoukelli2021multimodal, blip2}, fusion is performed across several layers, \eg, 24 for UNITER. However, for purely supervised multimodal tasks, such as affective understanding of human-centered video clips, the fusion mechanisms utilized are rather shallow. In particular, they usually involve combining pretrained architectures with shallow fusion mechanisms.

The focus of this work is on \emph{Multimodal Sentiment Analysis} (MSA). MSA involves understanding sentiments by interpreting behavioral signals such as speech, language, facial expressions, and body language~\cite{narayanan_2013_behavioral_signal}. 
Despite advancements in this area~\cite{metallinou2012context, zadeh2017tensor, sun2022learning}, the development of an architecture that employs a deep fusion scheme remains an open challenge.
The predominant body of research focuses on designing increasingly complex architectures~\cite{zadeh2017tensor, tsai20routing, zhang-etal-2023-learning-language} and training recipes~\cite{han2021improving, yu2021learning, sun2023efficient}, while efforts focusing on deep fusion schemes~\cite{georgiou2019deep, tsai2019multimodal, hu-etal-2022-unimse} remain restricted to fusion schemes involving three layers at most. Moreover, there is limited understanding of optimal capacity allocation for capturing multimodal interactions. In this work, we explore the questions of \emph{when does depth and scale help multimodal fusion} and \emph{how much dedicated capacity is optimal for processing multimodal information} in the context of MSA.

We propose \ours, a novel MSA fusion scheme that focuses on \textit{deepening the fusion process rather than complicating the fusion architecture}.
Our approach utilizes a pretrained language model (LM) and augments it with a small set of learnable fusion tokens appended after the language tokens.
These fusion tokens serve two key functions: 1) accumulate multimodal information, and 2) maintain linguistic and non-linguistic information flow through the network by design. At the same time, the textual modality is restricted from directly affecting other modalities, providing an inherent bias against language dominance.
The acoustic and visual cues are processed via a multimodal encoder, which injects multimodal information into the fusion tokens of the LM through novel cross-attention blocks. This design choice again allows for the accumulation of multimodal information solely in the fusion tokens while retaining the audiovisual information flow through the multimodal encoder.
The proposed fusion process can be repeated across multiple decoder LM layers, providing depth and scalability to our approach. The overall learning objective combines task losses for each individual modality and is coupled with language model regularization via language embedding augmentation and language modelling loss. 

What sets \ours apart is the deep and scalable fusion framework that, in synergy with the learnable fusion tokens and the training recipe, achieves enhanced fusion benefits. \ours by design allows for adjustable fusion depth and multimodal fusion token allocation. \ours emerges as a strong multimodal LM deep fusion architecture across MSA benchmark datasets. To the best of our knowledge, we are the first to explore both deep fusion configurations and multimodal capacity allocation in the MSA literature.

We validate the effectiveness of \ours through experiments conducted on three widely used MSA benchmark datasets: MOSI~\cite{zadeh2016multimodal}, MOSEI~\cite{zadeh2018multimodal}, and SIMS~\cite{yu2020ch}. These datasets were selected to cover different languages, dominance scenarios, and data availability conditions.
We compare our method with reproduced state-of-the-art fusion approaches~\cite{mao-etal-2022-sena} and provide a comprehensive analysis of the \ours components (fusion depth, fusion tokens, and encoder quality). Our contributions are:
\begin{enumerate}
    \item We introduce \ours, a novel multimodal language model with learnable tokens for deep fusion. \ours leverages pretrained LMs and augments them using a small set of learnable fusion tokens that progressively integrate multimodal information. Our design naturally promotes deep fusion schemes and allows for adjusting the number of fusion tokens for multimodal information processing.
    \item We propose a novel cross-attention fusion mechanism (MM Block) which captures interactions between the fusion tokens and the audiovisual information. Non-linguistic features are first processed by a dedicated encoder and then integrated with language representations at various depths through MM Blocks. This design maintains independent information flow for each modality in the network, allowing multimodal information to accumulate exclusively in the learnable tokens alone. 
    \item Through extensive experiments on MOSI, MOSEI and SIMS, we demonstrate that \ours achieves state-of-the-art performance across different languages and dataset scenarios. Our analysis reveals that deeper fusion schemes (5-7 layers), consistently outperform shallower approaches, and show that a small number of fusion tokens (8-20) achieves optimal results.
    \item We examine the importance of representation learning order through audiovisual encoder initializations, demonstrating the benefits of progressive multimodal representation learning. We also scale \ours up to small (1.7B) billion LLMs and provide comprehensive analysis on the interplay of fusion mechanisms, loss terms, and language embedding augmentation, illustrating the synergistic nature of our framework.
\end{enumerate}

The paper is structured as follows: \autoref{sec:related_work} covers related work and \autoref{sec:background} provides formulation and technical background for our study. Next, \autoref{sec:method} details the \ours architecture, and \autoref{sec:exp_setup} outlines our experimental setup. Our experimental analysis lies in~\autoref{sec:experiments} and starts with a comparison of \ours with reproducible state-of-the-art approaches (\autoref{sec:main_results}). An analysis on the interplay of performance, fusion depth, the number of fusion tokens, the audiovisual encoder intilization, and the impact of language distribution is provided in \autoref{sec:exp-analysis}. In \autoref{sec:exp_ablation} we present ablation results on various alternatives for the fusion mechanism. We also examine the importance of loss terms in the total objective, the impact of embedding regularization method, and gating mechanism variants. Finally, \autoref{sec:discussion} highlights method limitations, draws conclusions and discusses future research directions.

\section{Related Work}
\label{sec:related_work}

This section provides an overview of the literature, beginning with an exploration of works on multimodal fusion, which is the foundamental MML challenge. We then discuss advancements in the MSA field, which is the core of our experimentation.


\subsection{Multimodal Fusion}
In this section, we provide an analysis of multimodal fusion and, in particular, offer a dual perspective on fusion granularities. First, we outline the basic fusion mechanisms-operations (microscopic view) that may be employed for learning multimodal representations, and then, we discuss deep fusion schemes (macroscopic view) utilized across various tasks.

\paragraph*{Fusion Mechanisms}
Simpler fusion methods include addition, multiplication, and concatenation, followed by a projection step. Concatenation, in particular, is a popular choice among real-world supervised multimodal setups~\cite{afouras2018deep, yu2021learning}. More advanced techniques include tensor products~\cite{zadeh2017tensor} between latent modality representations and higher-order polynomial fusion~\cite{polynomial_fusion_19_nips}. Gating mechanisms are also widely used, with variations ranging from cell-like approaches such as GMU~\cite{gated_iclr_17}, to attention mechanisms~\cite{bahdanau2015neural, vaswani2017attention}. 
Specifically, fusion through attention comes in several flavors. Directional (non-symmetric) attention, such as \textit{cross-attention}~\cite{Lu2019ViLBERTPT, tan2019lxmert, alayrac2022flamingo, chen2023xllm}, fuses information from one modality\footnote{The term modality here is used loosely.}(acting as keys, values) to another (acting as queries). 
\textit{Causal attention}~\cite{tsimpoukelli2021multimodal, liu2023llava} is also non-symmetric and, processes multimodal input (token) information in an autoregressive manner. \textit{Self-attention}~\cite{chen2020uniter, li2020oscar}, processes multimodal information in a bidirectional manner and is considered a symmetric operation. 
Combinations of these mechanisms~\cite{alayrac2022flamingo, li2023blip} can also form larger fusion modules. Besides attention, other strategies employ graph neural networks (GNNs) to create nodes for each latent modality or entities such as speakers and objects, and fuse them via message passing across the constructed nodes~\cite{gao2020multi, MMGCN, Wang_2023_ICCV}.

\paragraph*{Deep Fusion Schemes}
Moving beyond early, late and hybrid schemes, the predominant paradigm in the deep learning era is deep fusion, \ie, fusion across multiple layers. 
ViLBERT~\cite{Lu2019ViLBERTPT} introduces the multimodal co-attention mechanism, applies it across several transformer layers, and shows that some tasks (or datasets) benefit from shallow (two-layer) while others benefit from deeper (six or more layers) fusion.
UNITER~\cite{chen2020uniter}, uses a single encoder transformer and appends visual and textual inputs. Fusion is performed via (bidirectional) self-attention modules across 24 layers.
Frozen~\cite{tsimpoukelli2021multimodal} feeds a pretrained large language model (LLM) with latent image representations followed by language tokens, and trains an image encoder while keeping the LLM frozen. Mutlimodal fusion is performed across the layers of LLMs via causal attention.
BLIP-2~\cite{blip2} and MiniGPT-4~\cite{zhu2024minigpt} follow similar, yet more sophisticated and better performing approaches than Frozen. 
LLaMA-Adapter~\cite{zhang2023llama} follows a different method and feeds the causal attention mechanism of the decoder with learnable layer-specific fused prompts, across the last layers of a frozen LLM. From a conditional image generation perspective, Unimo-G~\cite{li2024unimo} utilizes a multimodal transformer and conditions the layers of a diffusion model with multimodal information. For an extended review of multimodal LLMs and multimodal transformers, we refer to~\cite{yin2023survey} and~\cite{pami_survey_23}, respectively.

In a spirit more similar to our work, Ziegler \etal~\cite{ziegler2019encoder} propose an encoder-agnostic fusion and condition GPT-2~\cite{radford2019language} layers (12) for multimodal language generation. 
Flamingo~\cite{alayrac2022flamingo} inserts cross-modal information across the layers of LLMs, via gated cross-attention followed by a trainable and randomly initialized feed-forward block. Flamingo also feeds the decoder interleaved vision-language inputs. 
\ours advances this architectural direction with several key innovations. 
approach differs from Flamingo in several design choices.
We introduce a set of learnable fusion tokens that accumulate multimodal information progressively through the network. This design choice is closer to ideas from adapters~\cite{zhang2023llama}, perceiver~\cite{jaegle2021perceiver}, and bottleneck fusion~\cite{nagrani2021attention}. The appended set of learnable tokens gathers linguistic information via causal self-attention (CSA), while a parallel audiovisual encoder processes the non-linguistic signals. The fusion tokens are then integrated with the audiovisual information via cross-attention mechanisms across LM layers, offering fusion depth while maintaining independent information flow for each modality. Our novel gated cross-attention block improves upon the gated cross-attention mechanism of~\cite{alayrac2022flamingo,kongaudio, dubey2024llama3} by 1) initializing the feed-forward block from the corresponding LM block and further tuning it\footnote{Full fine-tuning and LoRA adapters are integrated in \ours}, 2) restricting cross-modal attention between non-linguistic and fusion tokens alone, to ensure efficient computation and allow for independent information flow, and 3) implementing sigmoid gating for optimal information flow control, as demonstrated in our ablations.

\subsection{Multimodal Sentiment Analysis}

MSA research mainly focuses on building better fusion schemes and utilizing diverse learning recipes to enhance representation learning for the task at hand. In particular, TFN~\cite{zadeh2017tensor} employs outer product of unimodal representations to capture cross-modal interactions. Poria \etal~\cite{poria2017multi} and Gu \etal~\cite{gu2018multimodal} implement multi-level and hierarchical attention to better contextualize information. 
DHF~\cite{georgiou2019deep} is the first deep fusion approach for MSA, and serves as an inspiration for this work. Its simple yet deep fusion design across three language levels demonstrates that fusion depth (macroscopic design) is more crucial for performance than complex fusion mechanisms (microscopic design).

Other types of neural structures employed in MSA include neural memory modules~\cite{zadeh2018multimodal}, top-down fusion~\cite{paraskevopoulos2022mmlatch}, capsule networks~\cite{tsai20routing}, and GNNs~\cite{joshi-etal-2022-cogmen}.
Tsai \etal~\cite{tsai2019multimodal} utilize transformers, where cross-attention blocks act as early fusion and concatenation serves as late fusion. Rahman \etal~\cite{rahman2020integrating} fine-tune a pre-trained BERT~\cite{devlin2019bert} model by incorporating a multimodal shifting layer as early fusion, and
Zhang \etal~\cite{zhang-etal-2023-learning-language} use language-guided fusion along with a fused hypermodality. 
In CENet~\cite{CENet_2023} authors exploit a pretrained language model tailored towards sentiment analysis instead of BERT.

Another line of work utilizes more complex learning recipes such as canonical correlation analysis~\cite{sun2022learning} and cycle-consistency loss~\cite{pham2019found} across modalities. Coupling different learning recipes with pre-trained models has been a popular choice among researchers. Yu \etal~\cite{yu2021learning} introduce a unimodal pseudo-labeling module that backpropagates three additional losses. Hazarika \etal~\cite{hazarika2020misa} augment the learning objective with feature reconstruction loss as well as attracting and repelling objectives. A two-step hierarchical learning recipe based on mutual information maximization is proposed in~\cite{han2021improving}, while Sun \etal~\cite{sun2022learning} propose a meta-learning framework that learns each unimodal network and then adapts them for the MSA task. Sun \etal~\cite{sun2023efficient} propose a transformer architecture leveraging dual-level reconstruction loss and an attraction loss in a Siamese setup between complete and incomplete data.
NIAT~\cite{niat_2024} learns a unified joint representation between clean and noisy data by coupling masking-based feature augmentation with an adversarial training strategy.
Hu \etal~\cite{hu-etal-2022-unimse} employ a text generation encoder-decoder architecture, using T5~\cite{RAFFEL2020T5}, and implement a contrastive loss among unimodal encoders. Fusion is performed across the last 3 layers of the transformer's encoder. 
The decoder generates text sequences, which are in turn decoded into MSA-related info such as polarity.
Notably, none of the aforementioned approaches utilizes more than three fusion layers.
\section{Background}
\label{sec:background}

\begin{figure}[!htb]
    \centering
    \includegraphics[width=\linewidth]{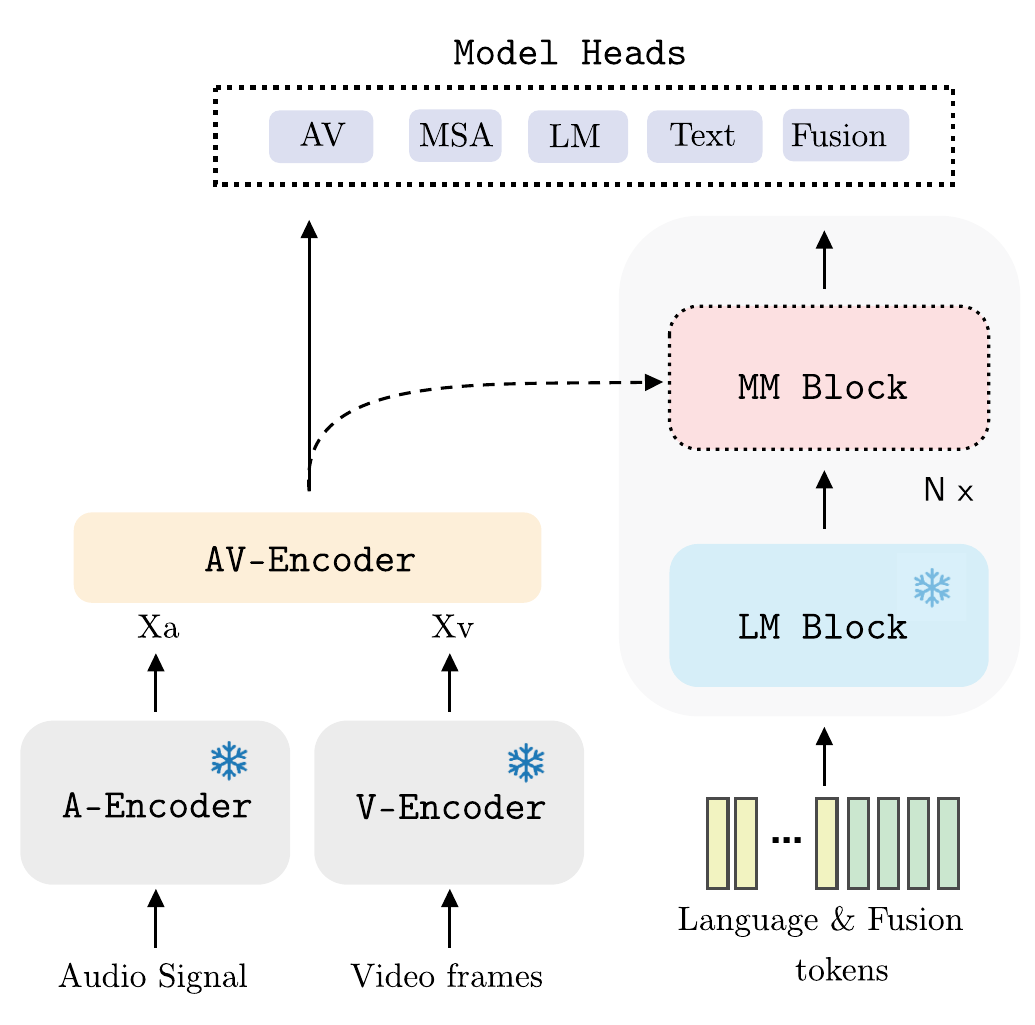}
    
    \caption{\emph{\ours architecture overview}. Audio and Visual features are being processed by a trainable \emph{AV-Encoder} and then fed to the Language Model (LM) for deep multimodal fusion. The LM consists of $N$ layers where the \emph{LM Block} remains frozen and the \emph{MM Block} is trainable. The output of the overall architecture are audiovisual, language and fused tokens which encapsulate audiovisual, linguistic and mutlimodal sentiment information respectively. The \emph{Model Heads} denote the involved objectives in out training recipe. 
    }
    \label{fig:msagpt}
\end{figure}

We first formulate \emph{multimodal sentiment analysis} (MSA) as a multimodal fusion task. 
We then present the transformer architecture with an abstractive notation suitable for the variants in this paper, and briefly outline the (conditional) language modelling objective.


\subsection{Problem formulation and notation}
Vectors and matrices are denoted by lowercase and uppercase bold letters respectively, \ie,
$\vx$ and $\vX$. Tensors are represented as $\tX$, and sets with calligraphic letters $\cM$. 
Depending on the context subscripts can denote timesteps ($\vx_t$) or modalities ($\vX_f$). Upperscripts in parenthesis depict different layers ($\vH^{(l)}$).
MSA is a task which takes as input three modalities, \ie, language, audio and video, and predicts the sentiment polarity. Each input modality $m$ resides in an input space $\cX_m \subseteq \mathbb{R}^{D_m \times L_m}$. Index $m$ denotes the modality from a set (of indices) $\cM = \{1,\dots, M\}$, $D_m$ is the input space dimensionality, and $L_m$ is the per modality (maximum) sequence length.
The multimodal input space can be expressed as the cartesian product of unimodal spaces $\cX_\cM = \cX_1 \times \dots \times \cX_M$.
Any supervised multimodal task can now be formulated as learning the (neural) mapping parameterized with $\vtheta$; $f_{\vtheta}: \cX_m \to \cU$, where $\cU \subset \mathbb{R}$ in the MSA case.
Each multimodal input ($m$-tuple) is represented as the collection of $M$ modalities as $\tX_i = [\vX_1, \cdots, \vX_M]$ along with a scalar label $y_i \in \cU$.
During the rest of the paper we denote the linguistic, the acoustic and the visual modalities with subscripts $t, a, v$ respectively. The audiovisual tokens are denoted with the $av$ subscript, and the (learnable) fusion tokens with the $f$ subscript.


\subsection{Transformer Architecture}
\label{sec:transformer}
Based on the transformer architecture paper~\cite{vaswani2017attention} we briefly outline its architecture and in particular the pre-norm Encoder-only and Decoder-only~\cite{decoder_2018} design, which are utilized across this paper.
Our presentation maintains a level of abstraction so that it can encapsulate transformer variants, and in particular different flavors in the attention mechanism~\cite{shazeer2019mqa}, the normalization and the feed-forward components~\cite{Touvron2023LLaMAOA}.

\subsubsection{Encoder Layer}
\label{sec:transf_enc}
The typical encoder layer design consists of a multihead Self-Attention (SA) module followed by a feed forward (FFW) block~\cite{vaswani2017attention}. We utilize the pre-norm transformer variant in our experiments.
Stacking encoder layers together, forms the Encoder transformer architecture.
Formally for an encoder layer $l$, and a latent input (from the previous layer) $\vH^{(l-1)}$ we have
\begin{align} \label{eq:transf_encoder}
	\wt{\vH}^{(l)} &= \vH^{(l-1)} + \SA(\Norm(\vH^{(l-1)})) \\
	\vH^{(l)} &= \wt{\vH}^{(l)} + \FFW(\Norm(\wt{\vH}^{(l)}))
\end{align}
where $\Norm$ denotes the normalization layer (LayerNorm, RMSNorm), $\SA$ the multihead self-attention layer, $\vH^{(l)}$ denotes the output representation of encoder layer $l$ (which is the input of layer $l+1$). All hidden representations $\vH^{(l-1)}, \vH^{(l)}$ lie in the same space $\mathbb{R}^{d \times L}$. The AV-Encoder of~\autoref{fig:msagpt} utilizes a stack of such Encoder Layers.

\subsubsection{Decoder Layer}
\label{sec:transf_dec}
The decoder follows a structure similar to the encoder with one main difference. Self-attention becomes causal self-attention (CSA), \ie, each position can only attend to its previous positions. Formally, given a hidden input representation $\vH^{(l-1)}$ to the $l$-th decoder layer the decoder layer equations are: 
\begin{align} \label{eq:transf_decoder}
	\wt{\vH}^{(l)} &= \vH^{(l-1)} + \CSA(\Norm(\vH^{(l-1)})) \\
	\vH^{(l)} &= \wt{\vH}^{(l)} + \FFW(\Norm(\wt{\vH}^{(l)}))
\end{align}
where $\CSA$ is the causal (masked) self-attention. The pretrained decoder LMs utilize a stack of such Decoder Layers, illustrated as LM Block in~\autoref{fig:msagpt}.


\subsection{Language Modelling Objective}
\label{sec:lm_loss}
Autoregressive or causal neural language models, generate text sequentially, predicting one token at a time, conditioned on previously generated/input tokens.
These LMs are trained based on minimizing the negative log-likelihood of the conditional probability $p_{LM}$ over a set of (previous) tokens $\vx_{<t}$ and some context information $\vZ$. Formally expressed:
\begin{align}
	L_{LM} = \sum_{t=1}^L -\log p_{LM}(\vx_t | \vx_{<t}, \vZ) \label{eq:lm_loss}
\end{align}
During this work the context information $\vZ$ denotes audiovisual (multimodal) information from the AV-Encoder (see~\autoref{fig:msagpt}).

\section{\ours}
\label{sec:method}

In this section, we describe the proposed multimodal language model framework. 
First, we introduce the novel multimodal language model architecture and describe its components, \ie, \emph{AV Encoder}, \emph{MLM}, and \emph{Model Heads}. Then we describe the training recipe, and finally include a discussion section on the components of \ours. 
\subsection{Overview}
\ours's main architectural components are the novel \emph{Multimodal LM (MLM)} and the \emph{AV-Encoder}. The \emph{MLM} consists of chained \emph{LM Block} and \emph{MM Block} modules, while \emph{AV-Encoder} is a standard encoder transformer architecture. The overall architecture is illustrated in~\autoref{fig:msagpt}.

The information flow through \ours is described below.
The \emph{A-Encoder} and \emph{V-Encoder} handle the feature extraction process (handcrafted or neural) for the acoustic and visual modalities, producing $\vX_a$ and $\vX_v$ features respectively. These extracted features are fed to \emph{AV-Encoder} which supplies the \emph{MLM}\footnote{should not be confused with MLM which is the masked language modelling, \eg, BERT-like approach~\cite{devlin2019bert}} with mutlimodal (audiovisual) information. In the MLM architecture, we append learnable fusion tokens $\vX_f$ after the language embeddings $\vX_t$ (~\autoref{fig:msagpt} assumes tokenization is already performed) and feed them to the MLM. The pretrained \emph{LM Block} processes the concatenated input and accumulates linguistic information at the learnable fusion tokens. After the (frozen) LM block, the fusion tokens alone interact with the audiovisual tokens within the \emph{MM Block}, and capture multimodal information. This process can be repeated for each decoder layer, rendering the proposed approach as a deep fusion scheme. Moreover, the audiovisual, linguistic, and fused components maintain their information flow through the network, and are fused in the \emph{Task Head} for the final prediction.
\subsection{AV Encoder}
The \emph{AV Encoder} 
fuses acoustic and visual information before feeding it in the MLM decoder.
We utilize two separate modality-specific transformer encoders to process unimodal information, and then fuse their representations through a feedforward network (Fusion FFW).
The modality-specific encoders process the (projected) acoustic and visual features, and output $\vZ_a$ and $\vZ_v$ respectively. These representations are concatenated (along the dimension-axis) and processed by the Fusion FFW, which outputs the fused $\vZ$. This process is summarized as:
\begin{equation}
    \vZ = \mathcal{E}_{AV}(\vX_{a}, \vX_{v}) \in \mathbb{R}^{d_{av} \times L}
\end{equation}
where $\vZ = \vZ_{av}$ is the multimodal information which is fed to the MLM through the AV Encoder. The AV Encoder is trained in isolation, and its weights are used to initialize the AV Encoder of \ours (see~\autoref{fig:msagpt}). 

\subsection{Multimodal Language Model (MLM)}
Here we describe the proposed Multimodal LM (MLM) architecture and focus on the learnable fusion tokens, the preatrained decoder \emph{LM Block}, and the novel \emph{MM Block} with its gated cross-attention ($\GCA$) and feed-forward projection ($\FFW$).

\subsubsection{Learnable Fusion Tokens}
We append $n_f$ learnable fusion tokens $\vX_f \in \mathbb{R}^{n_f \times d_t}$ to the pretrained LM input.
For any language input $\vX_{t} \in \mathbb{R}^{L_t \times d_t}$ (after tokenization and embedding layer) we have
\begin{equation}
    \vH^{(0)} = [\vX^{(0)}_{t} || \vX^{(0)}_{f}]
\end{equation}
where $[\cdot || \cdot]$ denotes concatenation, and $\vH^{(0)}$ is the input to the first LM Block. 
This set of fusion tokens appended after the language tokens serves two core functions: 1) gathers the linguistic information, and 2) interacts with the audiovisual information in the MM Block. 
The hyperparameter $n_f$ is analyzed in our experiments, and its optimal value is typically small, \ie, 12.

\subsubsection{LM Block}
Transformer based language modelling typically consists of stacked transformer decoder (see~\autoref{sec:transf_dec}) layers. 
Our approach differs in two ways compared to standard decoder LM: 1) LM Blocks process both fusion tokens and language tokens in every layer, and 2) all LM layers are kept frozen to minimize the number of trainable parameters and avoid catastrophic forgetting~\cite{ziegler2019encoder, chronopoulou-etal-2019-embarrassingly, alayrac2022flamingo, dubey2024llama3}.
Layer $l$ output goes to either the MM Block (when present) or the next LM Block ($l+1$).

\subsubsection{MM Block}
The proposed MM Block performs the core fusion operation through gated (multimodal) cross-attention ($\GCA$), followed by a feed-forward projection ($\FFW$).
MM Block takes the previous LM Block output $\wh{\vH}^{(l)} = [\wh{\vX}_t^{(l)} || \wh{\vX}_f^{(l)}]$, and the audiovisual context information $\vZ$ as input.
We split language and fusion tokens, and feed only fusion tokens to the $\GCA$ layer to capture multimodal interactions:
\begin{align}
    \wh{\vX}_t^{(l)}, \wh{\vX}_f^{(l)} &= \splt(\wh{\vH}^{(l)}) \\
    \wb{\vX}_f^{(l)} = \wh{\vX}_f^{(l)} &+ \sigma(a_1^{(l)}) \odot \GCA((\vZ), \Norm(\wh{\vX}_f^{(l)}))
\end{align}
The $\GCA$ mechanism utilizes audiovisual information $\vz$ as keys/values, and fusion tokens as queries. 
The fused tokens are then concatenated back with the language representation $\wh{\vX}_t^{(l)}$ and processed via the $\FFW$ layer as:
\begin{align} \label{eq:mm_block}
    \wb{\vH}^{(l)} &= [\wh{\vX}_t^{(l)} || \wb{\vX}_f^{(l)}] \\
    \vH^{(l)} &= \wb{\vH}^{(l)} + \sigma(a_2^{(l)}) \odot \FFW(\Norm(\wb{\vH}^{(l)}))
\end{align}
Output $\vH^{(l)}$ is fed to the next LM Block.
Here, $\sigma(\cdot)$ denotes the sigmoid gating function with learnable per-layer parameters $a_1^{(l)}, a_2^{(l)}$.
All hidden representations $\vH, \wh{\vH}, \wb{\vH}$ lie in $\mathbb{R}^{d \times L}$, with $L=L_t + n_f$.

The proposed scheme performs best when $\FFW$ is initialized with parameters from its corresponding LM Block $\FFW$ layer. We also experimented with other gating schemes, but found sigmoid most effective.

\begin{figure}[!htb]
    \centering
    \includegraphics[width=0.99\linewidth]{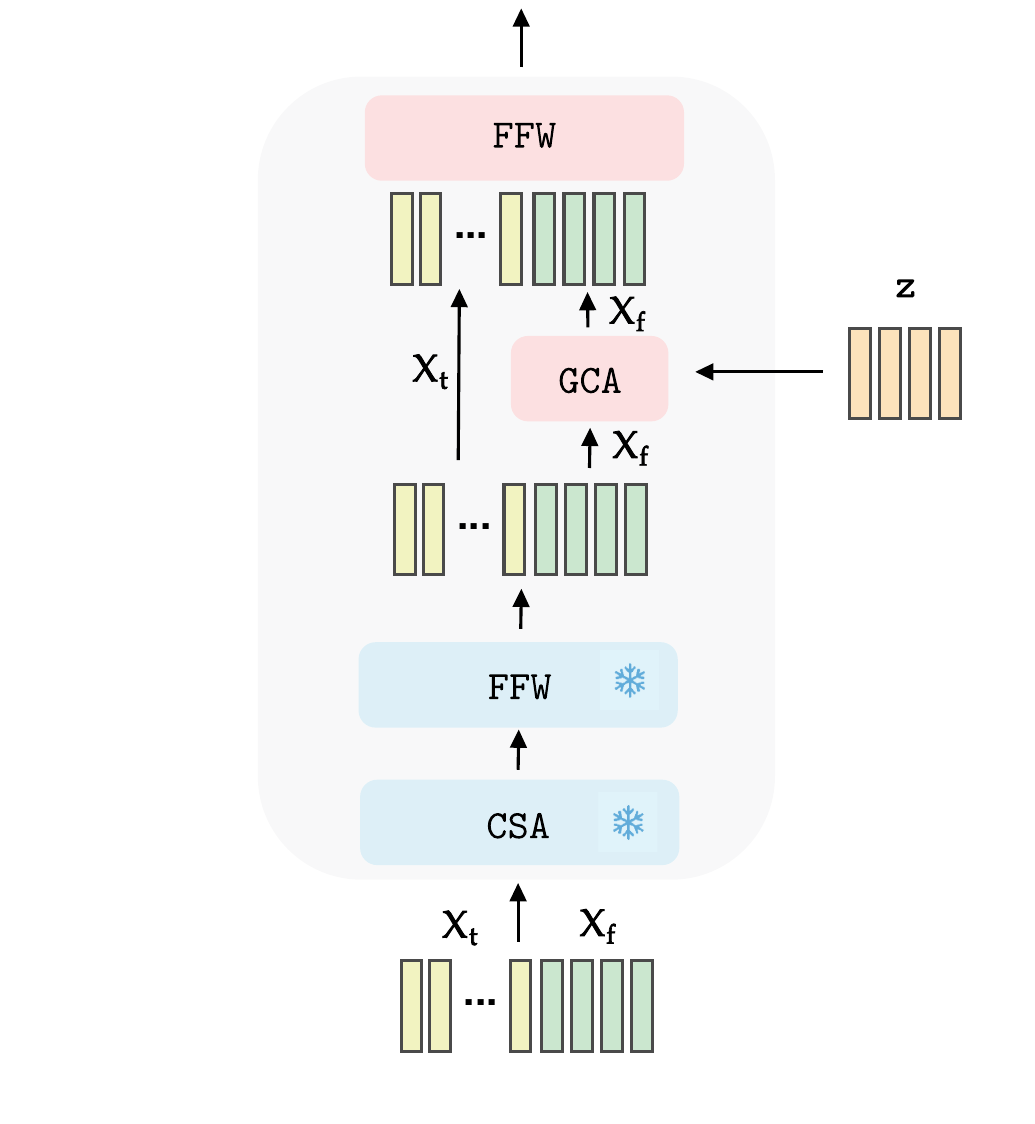}
    \caption{
    \emph{MM Block}. The fusion tokens (green) are appended at the LM input. First they accumulate linguistic information though the frozen LM Blocks ($\CSA$ and $\FFW$). Then the fusion tokens ($\vX_f$) are fed to the $\GCA$ module where they are fused with audiovisual information ($\vz$) and together with the language tokens ($\vX_t$) are fed to the $\FFW$ module of the MM Block. This modular design allows for integration of multiple MM Blocks across LM layers, enabling deep fusion capabilities.
    }
    \label{fig:mm_block}
\end{figure}


\subsection{Model Heads}

\paragraph*{Task Head}
A two layer MLP, $g(\cdot)$, operates as the task head, and maps the learned representations to the task space, \ie, sentiment polarity.
In particular, it accepts the pooled audiovisual representation $\langle\vz\rangle$, final layer's last ( assume position $K$) language embedding $\vX_{t}^{(N)}[K] = \vx_{t,K}$, and the mean fusion representation $\langle\vx_f\rangle = \langle\vX_f^{(N)}\rangle$:
\begin{equation}
    y_o = g([\langle\vz\rangle || \vx_{t,K} || \langle\vx_f\rangle])
\end{equation}
The mapping $g: \mathbb{R}^{2d+d_{av}} \to \cU$ is the late fusion operation.

\paragraph*{Modality Heads}
Linear mappings $W_{av}, W_{t}, W_{f}$ process audiovisual $\langle\vz\rangle$, textual $\vx_{t,K}$, and fused $\langle\vx_f\rangle$ representations for auxiliary task losses, producing sentiment polarity values $y_{av}, y_{t}, y_{f}$ respectively.

\paragraph*{Multimodal LM Head}
A linear layer $W_{\textsc{LM}}$ from the LM hidden space to the LM's vocabulary is adopted as in common language modelling. This layer is transfered from the pretrained LM (GPT2, SmolLM2) into our architecture and kept frozen during \ours's optimization process.


\subsection{Training Recipe}
This section presents the loss objectives and the regularization technique employed on language embeddings. We train \ours using $L_1(y, \hat{y}) = L_{\textsc{MAE}}(y, \hat{y}) = ||y - \hat{y}||$ modality specific loss terms, along with a LM loss.

\paragraph*{Task Loss}
The primary loss of our approach is $L_{msa} = L_1(y, y_o)$, based on the Task Head output.

\paragraph*{Auxiliary Loss}
For each modality, we also predict the ground truth multimodal polarity as an auxiliary loss term:
\begin{equation}
    L_{\text{aux}} = \lambda_{\text{av}}L_1(y, y_{\text{av}}) + \lambda_{t}L_1(y, y_t) + \lambda_{f}L_1(y, y_f)
\end{equation}
where $\lambda_{\text{av}}, \lambda_t, \lambda_{f}$ are weighting coefficients.

\paragraph*{Multimodal LM Loss}
Following benefits of LM loss in downstream tasks~\cite{chronopoulou-etal-2019-embarrassingly} to avoid overfitting and preserve language regularities, we employ (multimodal) LM loss $L_{LM}$ as in Eq.~\eqref{eq:lm_loss}. In \ours's case the conditioning information $\vc$ is the AV Encoder output $\vz$.

\paragraph*{Total Loss}
The total training objective is the combined task, auxiliary and reweighted multimodal LM loss.
\begin{equation}
    L_{\text{tot}} = L_{\text{msa}} + L_{\text{aux}} + \lambda_{\text{LM}}L_{\text{LM}} 
\end{equation}


\paragraph*{Language Embedding Regularization}
We apply regularization to the pretrained LM language embeddings, which aims at enhancing robustness against overfitting and language dominance~\cite{GKOUMAS2021WhatMakesTheDiff, georgiou21_interspeech}.
After tokenization, we employ SeqAug~\cite{georgiou2023seqaug} on the extracted language embeddings\footnote{\footnote{SeqAug is applied to the language embeddings alone and not to the fusion tokens}}.

\subsection{Implementation Details}
\label{sec:impl_details}
In this section we discuss the implementation details as well as key concepts of \ours.

\paragraph*{AV Initialization} We first pretrain the AV encoder separately, then fine-tune it when integrated with the LLM backbone. This approach aligns with other Multimodal LLM works that employ visual or acoustic backbones, and either fine-tune their encoder or use trainable connectors like Perceiver~\cite{jaegle2021perceiver, alayrac2022flamingo} or adapters ~\cite{hu2024wavllm}. For simplicity, we chose to fine-tune our AV encoder.

\paragraph*{Learnable Fusion Tokens} Our approach draws inspiration from methods like bottleneck fusion~\cite{nagrani2021attention}, but implements the concept differently. We append learnable tokens to the LM input and process them through two distinct steps: first using causal attention to gather linguistic information, then through cross-attention where they interact with multimodal data. This differs from bottleneck fusion's single step self-attention-only modality interaction. Therefore, \ours effectively extends the concept across both causal (decoder) and cross-attention layers.

\paragraph*{MM Block} Our Multimodal interaction block combines two components: $\GCA$ and $\FFW$. The $\GCA$ shares similarities with Flamingo, but differs in two key ways. First, we feed only learnable fusion tokens to cross-attention for more controlled fusion, while Flamingo uses the complete interleaved multimodal (language and vision) input. This reduces memory cost to $\mathcal{O}(n_{f}^2)$ compared to Flamingo's $\mathcal{O}((L_t + L_{\rm{v}})^2)$. Second, we use sigmoid gating instead of tanh, which proved more effective in our experiments. For the $\FFW$ component, we initialize it using parameters from its corresponding $\FFW$ LM Block. This initialization approach enhances performance without introducing training instabilities~\cite{alayrac2022flamingo}. We also integrate LoRA in MM Block $\FFW$ layer, to reduce training parameters and be robust against overfitting. This helps especially when using LLM backbones like SmolLM2.

\paragraph*{Auxiliary Task Losses} 
\ours maintains independent information flow for the audiovisual, language and fusion modalities by design. 
We add an auxiliary task loss for each one of these modalities.
This design choice follows evidence that better individual modality representations help better capture cross-modal interactions~\cite{du2023onunimodal} and improve representation predictability.

\paragraph*{Regularization}
To be more robust against overfitting and language dominance, we employ two regularization methods: multimodal LM loss and language embedding augmentation (via SeqAug). These techniques prove essential for typical language distributions, where the multimodal LM loss helps preserve language regularities~\cite{chronopoulou-etal-2019-embarrassingly} in the training data while preventing catastrophic forgetting, and SeqAug effectively samples from the underlying input language feature distribution~\cite{georgiou2023seqaug}. The effectiveness of these regularizers depends on the input data characteristics. Our experiments verify that for standard\footnote{in the sense that they could be similar to the pretraining corpus} language distributions, regularization is vital. However, for distributions significantly different from the pretraining corpus, regularization may not provide benefits. Therefore, \ours integrates both fusion and regularization components in its learning approach. Previous work in multimodal learning and MSA has either studied modality imbalance~\cite{wang2020makes, wu2022characterizing, huang_2021_mm_better, huang_22_mm_fail}, regularization~\cite{georgiou21_interspeech, liu2022_sims2, liu2023learning, georgiou2023powmix}, or robustness~\cite{hazarika2022analyzing, jin2023weakening}.
We are the first to propose an integrated multimodal fusion framework with regularization by design.
\section{Experimental setup}
\label{sec:exp_setup}

We evaluate \ours over three benchmark datasets for MSA to cover different languages, data abundance, and modality imbalance scenarios.

\subsection{Benchmark datasets}

\paragraph*{MOSEI}
The CMU-MOSEI~\cite{zadeh2018multimodal} dataset, is the largest MSA benchmark containing  $66$h of multimodal content. MOSEI offers a diverse range of samples containing $23,453$ manually transcribed and annotated utterance-level video segments from $1000$ distinct speakers, and covers $250$ topics. The average segment length is $7.28$ sec, with segmentation based on punctuation from the \emph{high-quality manual transcriptions}. Each segment is manually annotated in a Likert scale from -3 (strongly negative) to +3 (strongly positive). In our experiments, we observed a $21.98\%$ relative performance gap between text and the other modalities, with audio and visual features showing nearly identical performance levels.

\paragraph*{MOSI}
CMU-MOSI~\cite{zadeh2016multimodal} dataset contains approximately $2.5$h of YouTube videos (2-5 minute clips), consisting of 2,199 utterance-level movie review opinion segments from 93 videos and 89 different speakers (41 female, 48 male). Each segment averages 4.2 seconds and includes written transcripts and human annotator sentiment ratings on a likert scale from -3 to +3.

Compared to MOSEI, MOSI has key differences: it uses far fewer speakers (only about 10\%), covers a narrower range of topics (movie reviews), employs a smaller vocabulary size, and features more informal language (see also ~\autoref{tab:lang_factors}. Notably, MOSI exhibits the largest performance gap between text and audio (second best performing modality) features (33.94\%) among our experiments.

\paragraph*{SIMS}
CH-SIMS~\cite{yu2020ch} is a Chinese MSA becnhmark, with size comparable to MOSI, containing $2.3$h of $60$ high-quality videos, spanning movies, TV series, and variety shows. Researchers manually segmented the collected videos into $2,281$ utterance-level monologue video segments, averaging $3.67$ sec each. Human annotators transcribed the content, and assigned sentiment polarity scores ranging from -1 (strongly negative) to +1 (strongly positive).
SIMS is the most balanced MSA benchmark, showing relative modality performance gap $3.72\%$ in our experiments.


\subsection{Multimodal features}

Processing raw multimodal content
presents significant challenges, including high computational costs and potential copyright restrictions. Instead, researchers commonly use pre-extracted features that offer key advantages:
they reduce the heterogeneity gap between modalities (language, audio, video), and leverage information embedded into hand-crafted or neural representations. However, feature extraction pipelines in MSA vary across methods~\cite{mao-etal-2022-sena, lian2024merbench}, making comparisons across methods difficult. For consistency, we use the feature sets from Mao et.al.~\cite{mao-etal-2022-sena} across all methods and datasets.

\paragraph*{Text modality}

\ours is compatible with any decoder LLM model. For MOSEI we utilize the english pretrained \texttt{GPT2-large} model and its corresponding \texttt{GPT2-base} for the chinese\footnote{\url{https://huggingface.co/uer/gpt2-chinese-cluecorpussmall}} SIMS dataset. For MOSI we utilize a LLM, \ie, SmolLM2-1.7B\footnote{\url{https://huggingface.co/HuggingFaceTB/SmolLM2-1.7B}} to verify the scalability and compatibility of \ours. 
For other competitors and baseline models we follow Mao \etal~\cite{mao-etal-2022-sena} and use BERT~\cite{devlin2019bert} embeddings. In particular, we use \texttt{bert-base-uncased} for English and \texttt{bert-base-chinese} for the Chinese language. 

\paragraph*{Acoustic modality}
Acoustic analysis typically relies on extracted sound features. For MOSI and MOSEI datasets, researchers use COVAREP~\cite{degottex2014covarep} to extract 74 acoustic properties per frame, including pitch and 12 MFCCs. For SIMS, we use Librosa~\cite{mcfee2015librosa} to generate 33 acoustic features per frame.

\paragraph*{Video modality}
For video analysis, MSA tasks include facial landmarks, eye gaze, and facial action units.
MOSI and MOSEI use Facet\footnote{\url{https://imotions.com/platform}} to extract 35 facial action units linked to emotions and sentiment polarity. SIMS employs OpenFace2.0~\cite{baltrusaitis2018openface} to capture 709 features per frame, including 68 facial landmarks and 17 facial action units.


\subsection{Evaluation metrics}
\label{sec:eval_metrics}
We evaluate MSA as a regression task using \emph{mean absolute error} (MAE) and \emph{Pearson correlation} (Corr), following standard MSA practices~\cite{tsai2019factorized, hazarika2020misa, mao-etal-2022-sena}. We also map continuous sentiment predictions into discrete categories and measure classification accuracy (Acc-$k$). Our evaluation includes binary metrics (Acc-$2$ and $F1$), as well as $3$-class, $5$-class and $7$-class accuracies, depending on the benchmark requirements.


\subsection{Competitors and MSA models}
We evaluate our approach against leading MSA models using the M-SENA~\cite{mao-etal-2022-sena} framework for fair comparison. Our experiments include five state-of-the-art architectures: MulT~\cite{tsai2019multimodal}, MISA~\cite{hazarika2020misa}, Self-MM~\cite{yu2021learning}, ALMT~\cite{zhang-etal-2023-learning-language}, and TETFN~\cite{tetfn_2023}. These models have demonstrated strong performance across MSA datasets and provide a holistic performance overview. We also report other models performance from the literature.

\paragraph*{LF-DNN}
The \emph{late fusion deep neural network} (LF-DNN)~\cite{cambria2018benchmarking} processes each modality separately through neural networks before combining them for final prediction.

\paragraph*{TFN}
The \emph{tensor fusion network} (TFN)~\cite{zadeh2017tensor} uses LSTM for text processing while averaging acoustic and visual features. It fuses these processed features via Kroenecker product.

\paragraph*{MAG-BERT}
MAG-BERT~\cite{rahman2020integrating} enhances BERT with a multimodal adaptation gate to fuse information from audio and visual modalities.

\paragraph*{\textbf{MulT}}
The \emph{multimodal transformer} (MulT)~\cite{tsai2019multimodal} combines information across modalities through \emph{cross-attention} (CA) blocks. It then processes these fused representations using \emph{self-attention} (SA) mechanisms before concatenating for final prediction.

\paragraph*{\textbf{MISA}}
MISA~\cite{hazarika2020misa} processes audio and video using LSTM networks and fine-tunes BERT for text analysis. It embeds modalities into shared and modality-specific spaces to capture mutual information while preserving unique features. MISA fuses these representations in two ways: one branch reconstructs the input, while another uses \emph{self-attention} (SA) for the final multimodal prediction.

\paragraph*{\textbf{Self-MM}}
Self-MM~\cite{yu2021learning} uses LSTM networks for audio and visual processing and fine-tunes BERT for text. It utilizes \emph{unimodal label generation module} (ULGM) that creates individual modality labels from multimodal labels and embeddings. For prediction, Self-MM concatenates modality representations and processes them through dual linear layers. The model combines a main task loss with modality-specific losses from pseudolabeling.

\paragraph*{\textbf{ALMT}}
ALMT~\cite{zhang-etal-2023-learning-language} processes all modalities using transformers and fine-tunes BERT for text. It introduces an Adaptive Hyper-modality module that applies self-attention to language and cross-attention between text and other modalities. The model guides fused information via language-based cross-attention blocks.

\paragraph*{\textbf{TETFN}}
TETFN~\cite{tetfn_2023} applies text-based multi-head attention to enhance non-linguistic features with textual information. The model uses Vision-Transformer (ViT) for visual processing and combines cross-modal mappings with unimodal label prediction. All modalities are first encoded individually, then paired using text-oriented attention before final sentiment prediction.

\begin{table}
\centering
\caption{\emph{\ours configurations across datasets}. (L)LM: language backbone; MM Blocks: the LM layers after which we insert multimodal blocks; MM Layers: the total number of MM Blocks; $n_f$: the number of trainable fusion tokens; $\lambda_{*}$: loss weights; $\FFW$-FT: fine-tuning method for the $\FFW$ layer of each MM Block}
\begin{tabular}{lccc}
\toprule
 & \Th{MOSEI} & \Th{MOSI} & \Th{SIMS} \\
\midrule
(L)LM & GPT2-large & SmolLM2-1.7B & GPT2-base  \\
MM Blocks & 8-15-22-29-36 & 12-15-18-21-24 & 6:12 \\
MM Layers & 5 & 5 & 7 \\
$n_f$ & 12 & 8 & 16 \\
$\lambda_f$ & 1.0 & 0.4 & 1.0 \\
$\lambda_{\rm{av}}$ & 1.0 & 0.8 & 1.0 \\
$\lambda_t$ & 1.0 & 0.8 & 1.0 \\
$\lambda_{\rm{LM}}$ & 1.0 & 0.0 & 1.0 \\
$\FFW$-FT & Full & LoRA & Full \\
\bottomrule
\end{tabular}
\label{tab:model_configs}
\end{table}

\subsection{Implementation details}
We implement \ours in PyTorch~\cite{paszke2019pytorch}, 
using AdamW~\cite{loshchilov2018decoupled} optimizer with $\beta_1$=0.9, $\beta_2$=0.95, batch size $32$, warmup for one epoch, cosine annealing, and validation loss early stopping. For non-linguistic modalities (audio and video), we use the features provided in the M-SENA~\cite{mao-etal-2022-sena} framework. For language, MOSEI uses \texttt{GPT2-large}, SIMS the chinese \texttt{GPT2-base} and MOSI \texttt{SmolLM2-1.7B} LLM, with the latter employing LoRA~\cite{hu2022lora} adaptation ($r$= 512) in the MM Block's $\FFW$ layer. MM Blocks are inserted every $k$ layers backward from the final layer until performance plateaus. We tune $n_f$ in the range $\{8, 12, 16, 20\}$, set loss weights (auxiliary and MLM) to $1.0$ for MOSEI and SIMS, while tuning them in $\{0.0, \cdots , 1.0\}$ for MOSI, and adjust learning rates around 10$^{-4}$. For a detailed configuration we refer to \autoref{tab:model_configs}.
All baselines are reproduced using M-SENA and evaluation results are averaged over at least 5 independent runs. All experiments conducted in a single NVIDIA RTX 3090 (22GB).
\section{Experimental results}
\label{sec:experiments}

We evaluate and analyze \ours across multiple dimensions. First, we compare \ours against state-of-the-art reproduced algorithms from the literature. Our analysis then extends to algorithmic dimensions such as fusion depth, number of learnable fusion tokens, encoder initialization, language distribution characteristics, and scaling properties. The evaluation concludes with ablation experiments over components, such as loss terms, regularization strategies, and gating mechanism variants.

\begin{table}
\caption{\emph{Unimodal feature comparisons}. For each modality we evaluate the performance of pretrained models equipped with a trainable classification head. \textit{Native} denotes the features provided in the original papers and in~\cite{mao-etal-2022-sena}. For the language models we use different model weights for the english and the chinese languages. $\uparrow/\downarrow$: higher/lower is better. Red: worse than the baseline; bold: best for each MSA model.}
\label{tab:unimodal_performance}
\centering
\setlength{\tabcolsep}{3pt}
\begin{tabular}{lccccccccccccccccccccccccc}
\toprule
\mr{2}{\Th{Uni. Features}} & \mc{2}{MOSI} & \mc{2}{MOSEI} & \mc{2}{SIMS} \\ \cmidrule(lr){2-3} \cmidrule(lr){4-5} \cmidrule(lr){6-7}
& \Th{Acc2}$\uparrow$ &  \Th{Acc7}$\uparrow$ &
\Th{Acc2}$\uparrow$ & \Th{Acc7}$\uparrow$ &
\Th{Acc2}$\uparrow$ &  \Th{Acc5}$\uparrow$ &\\

\midrule
\textcolor{CB91_Blue}{\Th{\textbf{Language}}}   \\
BERT    & \bdark{78.96} & \bdark{31.83} & \bdark{83.39} & \bdark{50.31} & \bdark{77.35} & \bdark{37.71} \\
GPT2    & 66.85 & 24.30 & \blight{80.76} & \blight{48.71} & \bmedium{76.66} & \bmedium{34.61} \\
GPT2-large    & 73.45 & 29.88 & \bmedium{82.50} & \bmedium{49.82} & - & - \\
SmolLM2-1.7B$^*$    & \bmedium{75.96} & \bmedium{30.98} & - & - & - & - \\

\midrule
\textcolor{CB91_Amber}{\Th{\textbf{Audio}}} & 52.16 & 16.45 & 65.09 & 41.36 & 66.16 & 23.42 \\
\textcolor{my_Green}{\Th{\textbf{Vision}}}  & 43.09 & 15.5 & 64.41 & 41.88 & \bmedium{74.47} & \bmedium{25.46} \\
\bottomrule
\end{tabular}
\end{table}

\begin{table*}
\caption{\emph{State of the art comparisons for MOSI and MOSEI}. $^\dag$: results reported in~\cite{mao-etal-2022-sena};
$^*$: results reproduced; $\uparrow/\downarrow$: higher/lower is better. Blue: first and second best performing score per metric; bold: best for each MSA model.}
\label{tab:mosi_mosei_complete}
\centering
\begin{tabular}{lccccccccccccccccccccccccc}
\toprule
\mr{2}{\Th{Model}} & \mc{6}{MOSI} && \mc{6}{MOSEI} \\ \cmidrule(lr){2-7} \cmidrule(lr){9-14} 
& \Th{Acc2}$\uparrow$ & F1$\uparrow$ & MAE$\downarrow$ & \Th{Corr}$\uparrow$ & \Th{Acc5}$\uparrow$ &  Acc7$\uparrow$ && \Th{Acc2}$\uparrow$ & F1$\uparrow$ & MAE$\downarrow$ & \Th{Corr}$\uparrow$ & \Th{Acc5}$\uparrow$ &  Acc7$\uparrow$ \\

\midrule
LF-DNN$^\dag$    & 79.39 & 79.45 & 0.945 & 0.675 & - & -                   && 82.78 & 82.38 & 0.558 & 0.731 & - & -           \\
TFN$^\dag$    & 78.02 & 78.09 & 0.971 & 0.652 & - & -                   && 82.23 & 81.47 & 0.573 & 0.718 & - & - \\
MAG-BERT$^\dag$    & 83.41 & 83.47 & 0.761 & 0.772 & - & -                   && 84.87 & 84.85 & 0.539 & 0.764 & - & - \\
MulT$^*$ & 80.26 & 80.32 & 0.927 & 0.689 & 40.10 & 34.71                   && 84.07 & 83.93 & 0.564 & 0.731 & 53.97 & 52.56 \\
MISA$^*$    & 82.93 & 82.95 & 0.772     & 0.774     & 47.55 & 42.10           && 84.51 & 84.47 & 0.549 & 0.759 & 53.57 & 51.96 \\
TETFN$^*$ & \blight{84.10} & \blight{84.14} & \blight{0.725} & \blight{0.790} & \bmedium{52.77} & \bmedium{45.92} && \blight{85.20} & \blight{85.18} & 0.544 & 0.759 & \bmedium{55.54} & 53.74 \\
Self-MM$^*$ & \bmedium{84.22} & \bmedium{84.23} & \bmedium{0.724}     & \bmedium{0.791}     & \blight{52.22} & \blight{45.64}           && 84.26 & 84.24 & \bmedium{0.532} & \blight{0.765} & \blight{55.52} & \bmedium{53.85} \\
ALMT$^*$ & 83.90 & 83.89 & 0.746 & 0.784 & 48.77 & 43.58        && \bmedium{85.23} & \bmedium{85.32} & \blight{0.539} & \bmedium{0.766} & 54.64 & 53.05 \\
\midrule
\rowcolor{Gray3}
\ours    & \tb{85.60} & \tb{85.58} & \tb{0.692} & \tb{0.811} & \tb{53.18} & \tb{46.27}    && \tb{87.15} & \tb{87.10} & \tb{0.499} & \tb{0.804} & \tb{57.70} & \tb{55.88} \\

\bottomrule
\end{tabular}
\end{table*}

\subsection{Comparison with the state of the art}
\label{sec:main_results}

Our comparative analysis builds upon results of state-of-the-art models reproduced in the M-SENA framework~\cite{mao-etal-2022-sena}, utilizing their publicly available code and standardized feature sets. The performance results we obtained through reproduction align with those documented in M-SENA\footnote{\url{https://github.com/thuiar/MMSA/blob/master/results/result-stat.md}}.

\begin{table}
\caption{\emph{MOSEI} \ours performance for different sizes of GPT2 LM backbones.}
\label{tab:mosei_msalm}
\centering
\setlength{\tabcolsep}{4pt}
\begin{tabular}{lcccccc}
\toprule
\Th{\ours} & \Th{Acc2}$\uparrow$ & \Th{F1}$\uparrow$ & \Th{MAE}$\downarrow$ & \Th{Corr}$\uparrow$ & \Th{Acc5}$\uparrow$ & \Th{Acc7}$\uparrow$ \\
\midrule
base   & 85.83 & 85.74 & 0.529 & 0.773 & 56.61 & 54.65 \\
med & 87.00 & 86.98 & 0.511 & 0.795  & 56.89 & 55.01 \\
\rowcolor{Gray3}
large & 87.15 & 87.10 & 0.499 & 0.804 & 57.70 & 55.88 \\
\bottomrule
\end{tabular}
\end{table}

\paragraph*{MOSEI}
We train \ours with GPT2 backbones of various sizes on MOSEI. As illustrated in \autoref{tab:mosei_msalm}, utilizing the \texttt{GPT2-base} model as language backbone, achieves state-of-the-art performance across all metrics examined. Notably \autoref{tab:unimodal_performance} illustrates that despite generative LMs underperform embedding models as BERT, \ours outperforms all BERT-based competitors\footnote{This result align with literature findings~\cite{lian2024merbench}, where authors scale up to 13$B$ LLMs to match smaller embedding model performance.}, highlighting its efficacy as a fusion method. 
Since MOSEI is the larger MSA dataset, we scale \ours to \texttt{GPT2-medium} and \texttt{GPT2-large}, further improving the multimodal performance and pushing the limits of state-of-the-art for MOSEI as illustrated in \autoref{tab:mosi_mosei_complete} and \autoref{tab:mosei_msalm}. 
Overall, we get large improvements of 1.92$\%$ for Acc-$2$, 2.2$\%$ for Acc-$5$, and 2$\%$ for Acc-$7$, over the previous state-of-the-art models. We further discuss the role of scale and depth in \autoref{sec:exp-analysis}.

\paragraph*{MOSI}
For MOSI, GPT2 backbone performance significantly lags behind embedding models such as BERT (see~\autoref{tab:unimodal_performance}).
We therefore integrate SmolLM2-1.7B LLM into \ours, to get language-only performance closer to BERT. 
Despite the performance gap between our language backbone and BERT\footnote{We need to scale LLM further to get comparable performance~\cite{lian2024merbench}}, \ours achieves state-of-the-art multimodal performance on MOSI as illustrated in \autoref{tab:mosi_mosei_complete}. This result highlights that \ours is an efficient multimodal fusion method, since 1) it achieves a significant fusion improvement over the text-only performance, of 12.63\% relative Acc-$2$ improvement, compared to 6.65$\%$ relative improvement of Self-MM and TETFN, and 2) it is capable of operating fusion in scenarios with large modality performance gap (33.94$\%$). 
We present a detailed analysis on the MOSI case in \autoref{sec:exp-analysis}.

\begin{table}
\caption{\emph{State of the art comparisons for SIMS}. $^\dag$: results reported in~\cite{mao-etal-2022-sena};
$^*$: results reproduced; $\uparrow/\downarrow$: higher/lower is better. Blue: first and second best performing score per metric; bold: best for each MSA model.}
\label{tab:sims_complete}
\centering
\begin{tabular}{lcccccccr}
\toprule
\Th{Model}    & \Th{Acc}2$\uparrow$ & \Th{F1}$\uparrow$ & \Th{MAE}$\downarrow$   & \Th{Corr}$\uparrow$ \\
\midrule
LF-DNN $^\dag$      & 76.68 & 76.48 & 0.446  & 0.567 \\
TFN$^\dag$      & 77.07 & 76.94 & 0.437  & 0.582 \\
MAG-BERT$^\dag$      & 74.44 & 71.75 & 0.492  & 0.399 \\
MulT$^*$      & 78.56 & 78.66 & 0.453  & 0.564 \\
MISA$^*$      & 76.54 & 76.59 & 0.447  & 0.563 \\
TETFN$^*$      & \blight{79.21} & \blight{79.05} & \bmedium{0.419}  & \blight{0.592} \\
Self-MM$^*$      & \bmedium{80.04} & \bmedium{80.44} & 0.425  & \bmedium{0.595} \\
ALMT$^*$      & 78.16 & 78.16 & 0.433  & 0.575 \\
\midrule
\rowcolor{Gray3}
\ours & \tb{82.75} & \tb{83.15} & \tb{0.353} & \tb{0.729} \\
\bottomrule
\end{tabular}
\end{table}

\paragraph*{SIMS}
For SIMS, we utilize the Chinese GPT2-base model as the language backbone for \ours, though its performance is inferior to BERT (\autoref{tab:unimodal_performance}). Nevertheless, we achieve state-of-the-art results with significant relative improvements of 15.75$\%$ MAE, 22.52$\%$ Corr, and 3.38$\%$ Acc-$2$ over previous state-of-the-art multimodal approaches. This performance improvement demonstrates that \ours also achieves better fusion in balanced scenarios with narrower modality performance gaps.

\subsection{Analysis}
\label{sec:exp-analysis}

This section offers a detailed algorithmic analysis of \ours by studying factors such as fusion depth, number of learnable fusion tokens, encoder initialization, LM size and language data distributions. 


\subsubsection{The role of fusion depth and size}
We conduct experiments on both SIMS and MOSEI in this section. SIMS based experiments cover different fusion depth configurations for \texttt{GPT2-base} LM backbone, while MOSEI investigates the role of scaling the backbone itself, \ie, from \texttt{GPT2-base} to \texttt{GPT2-large}.

\paragraph*{Multimodal Fusion Depth} ~\autoref{tab:sims_depth} illustrates three different fusion setups for the SIMS dataset. The results clearly highlight the existence of an optimal fusion depth configuration, specifically seven MM Blocks in our case. This depth exceeds typical fusion depths from the literature which are primarily limited to three layers. Moreover, we observe that increasing depth beyond this optimal value slightly decreases performance, while decreasing depth significantly harms our results. This finding clearly demonstrates that adequate depth is necessary for efficient fusion.

Across all datasets and backbones, we observe that \emph{it is better to skip adding MM Blocks at the shallow LM layers}. We hypothesize that low-level linguistic features learned in these layers, do not effectively integrate with non-linguistic (audio-visual) signals.

\begin{table}
\caption{\emph{\ours fusion depth analysis on SIMS}. GPT2: language backbone size; MM Blocks (\#): the LM layers after which we insert multimodal blocks (their total number); MM Params: the number of total (mutlimodal) trainable parameters.}
\label{tab:sims_depth}
\centering
\setlength{\tabcolsep}{2.2pt}
\begin{tabular}{lcccccc}
\toprule
\Th{GPT2} & \Th{MM Blocks (\#)} & \Th{MM Params(M)} & \Th{MAE}($\downarrow$) & \Th{Corr}($\uparrow$) \\ 
\midrule
base   & 8:12 (5) & 29.75        & 0.374  & 0.697            \\
\rowcolor{Gray3-soft}
base   & 6:12 (7) & 41.65        & 0.353  & 0.729            \\
base   & 4:12 (9) & 53.55        & 0.358  & 0.724            \\
\bottomrule
\end{tabular}
\end{table}

\paragraph*{Fusion Scheme Size vs Depth} 
~\autoref{tab:mosei_depth} illustrates different fusion configurations for the MOSEI dataset, utilizing three progressively larger MM Block variants (and consequently larger GPT2 backbones). We observe a linear decrease (7, 6, 5) in the optimal fusion depth as the size of the MM Blocks increases. Our optimal configuration employs a fusion scheme of five layers, maintaining deeper fusion than competitor approaches, demonstrating the benefits of deep fusion.
Furthermore, our best performing model utilizes fewer trainable parameters than ALMT and Self-MM, further highlighting the efficiency of \ours.
The last two columns of~\autoref{tab:mosei_depth} display the relative (absolute) improvement compared to the base model. 
Consistent with literature, we observe decaying performance improvements (smaller deltas) when using larger LM backbones in MOSEI. This preludes a performance saturation with the increase of trainable parameters, similar to the one observed in SIMS~\autoref{tab:sims_depth}. For a fixed amount of data, there exists an optimal fusion depth, that achieves the best multimodal performance, typically exceeding that of existing competitors.

\begin{table}
\caption{\emph{\ours fusion depth analysis on MOSEI}. GPT2: language backbone size; MM Blocks (\#): the LM layers after which we insert multimodal blocks (their total number); MM Par.: the number of total (mutlimodal) trainable parameters in millions (M); $\Delta$(Metric): relative metric improvement ($\%$) from base \ours model.}
\label{tab:mosei_depth}
\centering
\resizebox{\linewidth}{!}{
\begin{tabular}{lcccccc}
\toprule
\Th{GPT2} & \Th{MM Blocks(\#)} & \Th{MM Par.} & $|\Delta$\Th{Acc}2$|$ & $|\Delta$\Th{MAE}$|$ \\ 
\midrule
base      &4-6-8:12 (7) & 41.65        & -- & --            \\
med.      &9-12-15-18-21-24 (6) & 63.30          & 1.36 & 3.40                \\
\rowcolor{Gray3}
large  & 8-15-22-29-36 (5) & 82.35          & 0.17& 2.35               \\
\bottomrule
\end{tabular}
}
\end{table}


\subsubsection{The impact of learnable fusion tokens ($n_f$)}
The optimal number of learnable fusion tokens ($n_f$) remains consistently small across all architectural configurations, with values varying according to modality performance gaps in different datasets. Specifically, we observe larger $n_f$ values for datasets with smaller modality performance gaps: SIMS requires 20 tokens, MOSEI 12, and MOSI 8 tokens respectively. 
\emph{This pattern suggests that datasets with more balanced modality contributions benefit from additional fusion tokens which capture richer multimodal interactions}.

\begin{figure}
\centering
\begin{subfigure}[h]{0.9\columnwidth}
\centering
\begin{tikzpicture}
\begin{axis}[
	font=\scriptsize,
	xlabel={$n_f$},
	ylabel={Acc-$2$ ($\uparrow$)},
	xmin=2, xmax=26,
	ymin=84, ymax=87.5, 
	xtick={4, 8, 12, 16, 20, 24},
	legend pos=south east, 
	height=.5\columnwidth,
	ymajorgrids=true,
]

\addplot[color=my_Green, mark=*] coordinates {
	(4, 84.31) 
	(8, 85.49)
	(12, 85.83)
	(16, 85.76)
	(20, 85.40)
	(24, 85.28)
};

\addplot[color=CB91_Violet, mark=*] coordinates {
	(4, 86.50)
	(8, 86.81)
	(12, 87.15)
	(16, 86.95)
	(20, 86.43)
	(24, 86.05) 
};

\end{axis}
\end{tikzpicture}

\label{fig:selfmm_mosei_ablation_acc2}
\end{subfigure}
\hfill
\begin{subfigure}[h]{0.9\columnwidth}
\centering
\begin{tikzpicture}
\begin{axis}[
	font=\scriptsize,
	xlabel={$n_f$},
        xmin=2, xmax=26,
	ymin=49.30, ymax=54.00, 
	xtick={4, 8, 12, 16, 20, 24},
	ylabel={MAE ($\downarrow$)},
	ytick={50.00, 51.00, 52.0, 53.00, 54.0},
	legend pos=north east, 
	height=.5\columnwidth,
	ymajorgrids=true,
        legend to name=myfancyname
]

\addplot[color=my_Green, mark=*] coordinates {
	(4, 53.46) 
	(8, 53.26)
	(12, 52.94)
	(16, 52.95)
	(20, 53.20)
	(24, 53.28)
};
\addlegendentry{\ours (base)}

\addplot[color=CB91_Violet, mark=*] coordinates {
	(4, 50.80) 
	(8, 50.50)
	(12, 49.96)
	(16, 50.30)
	(20, 50.77)
	(24, 50.99) 
};
\addlegendentry{\ours (large)}

\end{axis}
\node[anchor=north] at (current axis.below south) {\scriptsize{\ref*{myfancyname}}};
\end{tikzpicture}
\label{fig:selfmm_mosei_ablation_mae}
\end{subfigure}
\caption{Impact of the number $n_f$ of \emph{learnable fusion tokens} on MOSEI. $\uparrow/\downarrow$: higher/lower is better.}
\label{fig:nf_ablation}
\end{figure}
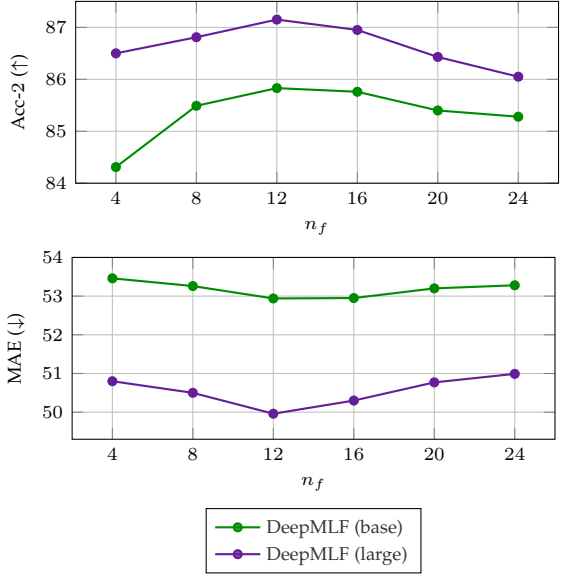

~\autoref{fig:nf_ablation} illustrates that $n_f$ transfers across LM backbones, showing consistent behavior as a robust hyperparameter. Increasing $n_f$ beyond its optimal value leads to performance degradation, suggesting that a small set of fusion tokens provides the most effective approach for multimodal information integration.

\subsubsection{The impact of encoder initialization}
In this experiment we evaluate the impact of the audiovisual encoder initialization scheme on the overall performance of \ours. We conduct experiments on both MOSEI and SIMS datasets, to assess the effect in language-dominated and more balanced setups. We examine three initialization strategies: pretrained\footnote{Pretraining of AV Encoder is performed on each MSA dataset before integrating into \ours.} encoder and further fine-tuned (\emph{Pre\&Tune}), frozen pretrained encoder (\emph{Pre\&Fro}), and randomly initialized encoder trained from scratch (\emph{Random\&Tune}).

The \emph{Pre\&Fro} setup shows a minor performance degradation compared to \emph{Pre\&Tune} for MOSEI and a larger for SIMS. This result shows that further tuning of the audiovisual encoder impacts \ours positively.
The \emph{Random\&Tune} configuration for MOSEI yields a larger performance degradation compared to \emph{Pre\&Fro}. SIMS exhibits a substantial performance degradation under random initialization and training from scratch.

We attribute these outcomes to each dataset's modality characteristics. In the smaller and more balanced SIMS dataset, audiovisual features contribute significantly to multimodal performance, making encoder initialization a crucial factor for achieving optimal performance. For MOSEI, which is more language dominated, the AV Encoder effectively captures task-relevant audiovisual features, resulting in more robust performance across initialization schemes.

These results highlight the importance of a progressive learning approach (fusion curriculum), \ie, first developing unimodal representations (pretrained LM), then building audiovisual representations (AV encoder) and finally learning multimodal connections, rather than jointly learning everything from scratch. This finding on fusion curriculum aligns with broader multimodal learning literature~\cite{du2023onunimodal} suggesting that progressive representation acquisition benefits multimodal representation learning.

\begin{table}
\caption{Assesing the impact of encoder initialization in \ours. \emph{Pre\&Tune}: initialize encoder from pretrained and further tune; \emph{Pre\&Fro}: intitialize encoder from pretrained and keep frozen; \emph{Random\&Tune}: initialize encoder randomly and tune it.}
\label{tab:encoder_quality}
\centering
\begin{tabular}{lccccccc}
\toprule
\mr{2}{\Th{Setting}} & \mc{2}{MOSEI} && \mc{2}{SIMS} \\ 
\cmidrule(lr){2-3} \cmidrule(lr){5-6} 
& \Th{Acc2}$\uparrow$ & MAE$\downarrow$ && \Th{Acc2}$\uparrow$ & MAE$\downarrow$ \\
\midrule
\rowcolor{Gray3}
\emph{Pre\&Tune} & 87.15 & 0.499 && 82.75 & 0.353 \\
\emph{Pre\&Fro} & 86.94 & 0.504 && 81.99 & 0.363 \\
\emph{Random\&Tune} & 86.59 & 0.508 && 80.43 & 0.382 \\
\bottomrule
\end{tabular}
\end{table}


\subsubsection{The impact of language data on \ours}

\ours demonstrates effective performance across both English (MOSI, MOSEI) and Chinese (SIMS) languages. This cross-lingual capability, while desirable in language fusion algorithms, is not uniformly observed in MSA models. Notably, BERT-Chinese features alone achieves 77.35 Acc-$2$ (~\autoref{tab:unimodal_performance}), surpassing several multimodal fusion approaches shown in \autoref{tab:sims_complete} and documented by Mao \etal~\cite{mao-etal-2022-sena} (Table 4).

\begin{table}
\caption{Language characteristics for MOSI and MOSEI. These factors highlight key differences in language distributions between the two datasets. The final row demonstrates SBERT's ability to discriminate between sentences from each distribution.}
\label{tab:lang_factors}
\centering
\begin{tabular}{lcccccc}
\toprule
\Th{Factors} && \Th{MOSI} & \Th{MOSEI} \\ 
\midrule
Sentences  && 2,199   & 23,453     \\
Vocabulary Size && 3,107 & 23,026         \\
Sentence Length(s)    && 4.2  & 7.3   \\
Topics && Movie Reviews & Diverse \\
Tone && Informal & Mixed \\
\midrule
\midrule
\Th{MOSI} vs \Th{MOSEI} && \Th{Acc2} & 90.0(\%) \\
\bottomrule
\end{tabular}
\end{table}

We focus our analysis on the MOSI dataset, where we need to scale up to billion parameter LLMs (SmolLM2-1.7B) to achieve performance comparable, though still inferior, to much smaller embedding models. To understand this behavior, we compare MOSI with MOSEI, both English datasets, by first examining the impact of dataset size. We conduct experiments on MOSEI by randomly sampling 300 subsets, each containing 5-10\% of the full data, training a classifier on GPT2-base embeddings, and evaluating on MOSEI's test set. GPT2-base achieves 77\% Acc-$2$ on these MOSEI subsets on average, matching SmolLM2-1.7B's MOSI performance but exceeding GPT2-base's MOSI results by 10\%. This GPT2 performance gap between MOSI and MOSEI subsets suggests additional factors affecting MOSI's language distribution, besides size.

These factors, illustrated in~\autoref{tab:lang_factors}, demonstrate that MOSI's language is characterized as informal and topic-specific. We hypothesize that these factors create a \emph{language distribution shift} compared to GPT2's training dataset, which contains high-quality web pages that have been curated/filtered by humans~\cite{radford2019language}. This distribution shift explains why GPT2s show degraded performance on MOSI, and why \ours does not improve with the addition of the LM task loss. To validate our hypothesis we utilize SBERT~\cite{reimers-2019-sentence-bert} embeddings\footnote{\url{https://huggingface.co/sentence-transformers/all-mpnet-base-v2}} and train a linear classifier to distinguish between MOSI and MOSEI language samples. This classifier achieves 90\% binary accuracy, confirming that the two language distributions are separable, \ie, the distribution shift exists.

\paragraph*{Findings} \ours shows the desirable property of multilingual capability. Furthermore, \ours leverages (L)LM backbones that are primarily designed for generation rather than downstream tasks, making it more sensitive to language distribution shifts. To address this, in cases like MOSI, we utilize larger LLM backbones, which better capture language characteristics. Notably, integrating small-billion LLMs (1.7B) enables \ours to outperform all encoder-based (\eg BERT) competitors in multimodal scenarios.

\subsection{Ablation study}
\label{sec:exp_ablation}

\subsubsection{Fusion Scheme Ablation}
In this experiment, we systematically evaluate different configurations of our proposed fusion mechanism. As described in~\autoref{eq:mm_block}, the MM Block consists of two primary modules: the Gated Cross-Attention ($\GCA$) and the Feed-Forward Network ($\FFW$). We modify the $\GCA$ module to investigate all four possible input combinations of fusion $\vX_f$ and language tokens $\vX_t$. Additionally, we examine the impact of the $\FFW$ layer by conducting experiments where this component is omitted from the architecture. Our experiments are carried on SIMS. 

\ours setup, where fusion tokens $\vX_f$ interact with audiovisual features $\vZ$ followed by a $\FFW$ layer, achieves the best results. Further injecting linguistic information $\vX_t$ into the $\GCA$ mechanism slightly degrades performance (second row). In the third row, when we omit the fusion tokens from the $\GCA$ and retain only the language tokens, we observe a larger performance drop. Finally, removing the $\FFW$ from our setup significantly impacts performance negatively.

These results highlight two crucial factors in the fusion mechanism design. First, the $\FFW$ component plays an essential role in the overall fusion process. Second, the design decision to implement controlled interaction specifically between fusion tokens and audiovisual tokens, without additional elements, produces optimal performance.

\begin{table}
\caption{\emph{Fusion block ablation}. The fusion mechanism consists of two main parts. The $\GCA$ (gated cross-attention) which is a multimodal cross-attention operation between the audiovisual information $\vZ$ and a combination of fusion ($\vX_f$) or language ($\vX_t$) tokens. The $\FFW$ which is inserted on top of the $\GCA$ following standard encoder-decoder design. The colored line illustrates the scheme utilized in \ours. $\uparrow/\downarrow$: higher/lower is better.}
\label{tab:fusion_ablation}
\centering
\begin{tabular}{lcccccccr}
\toprule
$\GCA$ & $\FFW$ && \Th{Acc}2$\uparrow$ & \Th{F1}$\uparrow$ & \Th{MAE}$\downarrow$   & \Th{Corr}$\uparrow$ \\
\midrule \rowcolor{Gray3}
$(\vX_f, \vZ)$ & \checkmark     && 82.75 & 83.15 & 0.353  & 0.729 \\
$(\vX_t || \vX_f, \vZ)$ & \checkmark && 82.20 & 82.49 & 0.359  & 0.721 \\
$(\vX_t, \vZ)$ & \checkmark     && 81.14 & 81.46 & 0.365  & 0.711 \\
$(\vX_f, \vZ)$ & \ding{55} && 80.88 & 81.32 & 0.384  & 0.678 \\
\bottomrule
\end{tabular}
\end{table}


\subsubsection{Loss term ablation}
Our ablation study in~\autoref{tab:ablation_mixing_config} reveals a consistent pattern in the importance of loss terms across both MOSEI and SIMS datasets, consisting of a primary and a secondary tier.
The primary tier consists of the fusion loss $L_f$ and language modelling loss $L_{\rm{LM}}$, which emerge as the most crucial terms of our objective function. Removing them leads to the largest performance drops (MOSEI: 0.73/0.44 Acc-$2$, and SIMS: 1.57/1.61 Acc-$2$, for $L_{f}$ and $L_{\rm{LM}}$). The fusion loss $L_f$ plays a fundamental role by forcing the network to accumulate meaningful task-predictive multimodal information in the set of learnable fusion tokens. Meanwhile, the LM loss serves as a critical regularization mechanism, preventing catastrophic forgetting on the pretrained LM.

In the secondary tier our ablation places the audiovisual ($L_{\rm{av}}$) and language($L_{t}$) loss terms. Specifically, for both datasets, $L_{t}$ and $L_{av}$ have comparable effects, \ie, 0.79 vs 1.09 Acc-$2$ for SIMS, and 0.42 vs 0.26 Acc-$2$ for MOSEI. Moreover, MOSEI-traned \ours appears to be more robust against loss term removal, exhibiting smaller performance drops compared to SIMS-trained \ours.
\begin{table}
\caption{\emph{Loss term ablation on \ours}: audiovisual loss ($L_{\rm{av}}$), text loss ($L_{t}$), Fusion Token Loss ($L_f$), and causal language modelling loss ($L_{\rm{LM}}$). The table illustrates results as MOSEI/SIMS for both Acc-$2$ and MAE. $\uparrow/\downarrow$: higher/lower is better; \colorbox{yellow}{Tier-1}, \colorbox{green}{Tier-2}}
\label{tab:ablation_mixing_config}
\centering
\begin{tabular}{ccccccccc}
\toprule
\colorbox{green}{{$L_{\rm{av}}$}} & \colorbox{green}{$L_{\rm{t}}$} & \colorbox{yellow}{$L_{\rm{f}}$} & \colorbox{yellow}{$L_{\rm{LM}}$} & \Th{Acc}-$2\uparrow$  
& MAE$\downarrow$ \\
\midrule \rowcolor{Gray3}
\ch & \ch & \ch & \ch & \tb{87.15}/\tb{82.75} & \tb{0.499}/\tb{0.353}  \\
    & \ch & \ch & \ch & 86.89/81.96 & 0.502/0.358 \\
\ch &     & \ch & \ch & 86.73/81.66 & 0.505/0.355  \\
\ch & \ch &     & \ch & 86.42/81.18 & 0.511/0.367 \\
\ch & \ch & \ch  &  & 86.71/81.14 & 0.512/0.377 \\
\bottomrule
\end{tabular}
\end{table}

\subsubsection{Augmentation Ablation}
In this ablation, we demonstrate the critical role of language embedding augmentation in \ours's performance by comparing our integrated SeqAug method against conventional approaches such as dropout and noise injection on the SIMS dataset. To ensure fair comparison, we independently tune dropout and noise injection hyperparameters and report the average scores of their best performing configuration in ~\autoref{tab:augmentation_ablation}. Our experiments reveal that embedding regularization is crucial for \ours's training recipe, with SeqAug emerging as significantly more effective than both alternatives (performance drops of 1.36 Acc-$2$ and 0.1 MAE for noise injection, 1.67 Acc-$2$ and 0.13 MAE for noise injection). 
SeqAug proves more suitable for our pretrained LM backbone, since it resamples from the underlying language distribution, \ie, performs a soft permutation in the sequence which acts as an augmentation while preserving semantics. Replacing SeqAug with alternative augmentation methods leads to performance degradation comparable to removing a tier-1 loss term from the total objective (recall drops of ~1.32 Acc-$2$ and ~0.2 for tier-1 losses). These results underscore that proper regularization through SeqAug is crucial for \ours's optimal performance.

\begin{table}
\caption{\emph{Language embedding augmentation ablation}. $\uparrow/\downarrow$: higher/lower is better.}
\label{tab:augmentation_ablation}
\centering
\begin{tabular}{lcccccccr}
\toprule
\Th{Embedding Aug.}    & \Th{Acc}2$\uparrow$ & \Th{F1}$\uparrow$ & \Th{MAE}$\downarrow$   & \Th{Corr}$\uparrow$ \\
\midrule \rowcolor{Gray3}
SeqAug      & 82.75 & 83.15 & 0.353  & 0.729 \\
Noise Injection      & 81.05 & 81.44 & 0.368  & 0.708 \\
Dropout      & 80.74 & 81.12 & 0.372  & 0.702 \\
\bottomrule
\end{tabular}
\end{table}

\subsubsection{Gating Mechanism Ablation}
In this ablation study, we examine variants of the gating mechanism, with analysis conducted on the MOSEI dataset. We compare our proposed sigmoid gating against two alternatives: 1) tanh gating (as used in Flamingo~\cite{alayrac2022flamingo}) and 2) complete gate removal (None), which results in a vanilla encoder-decoder architecture~\cite{vaswani2017attention}.

Our experiments demonstrate that sigmoid gating consistently outperforms both alternative approaches. 
Furthermore, the presence of any gating mechanism proves superior to non-gating configurations, though even the configuration without gating still achieves state-of-the-art performance. The primary differences between our proposed sigmoid and tanh gating are: 1) initialization strategy (sigmoid initialized at 0.5 versus tanh at zero/closed gate), and 2) gate bounds (sigmoid's bounded positive range versus tanh's $[-1,1]$ range).
The balanced initialization and positive range characteristics of sigmoid gating prove most effective for our MM Block implementation.

\begin{table}
\caption{\emph{Gating mechanism ablation}. Gating Mech: the gating mechanism studied; Init: the initial gate value; $\uparrow/\downarrow$: higher/lower is better.}
\label{tab:gating_ablation}
\setlength{\tabcolsep}{4pt}
\centering
\begin{tabular}{lcccccccr}
\toprule
\Th{Gating} & \Th{Init.} & \Th{Range}    & \Th{Acc}2$\uparrow$ & \Th{F}1$\uparrow$ & \Th{MAE}$\downarrow$ & \Th{Corr}$\uparrow$  \\

\midrule \rowcolor{Gray3}
sigmoid & 0.5 & $[$0,1$]$ & 87.15 & 87.10 & 0.499 & 0.804 \\
tanh & 0.0  & $[$-1,1$]$ & 87.01 & 86.98 & 0.505 & 0.800 \\
None & 1.0  & - & 86.86 & 86.84 & 0.511 & 0.798 \\

\bottomrule
\end{tabular}
\end{table}
\section{Discussion}
\label{sec:discussion}

\subsection{Limitations}
\label{sec:limitations}
\ours leverages pretrained LMs as language backbones, inheriting certain limitations of these decoder architectures. First, these models are primarily designed for generation rather than predictive tasks, which, as shown in our language distribution analysis, can impact predictive performance, and requires larger LM integration. Second, the autoregressive nature of language modelling has increased inference time compared to encoder-based (non-autoregressive) approaches. However, all existing literature regarding speeding up inference is directly transferable to \ours.

\subsection{Conclusion}
This work introduces \ours, a multimodal language model framework with learnable tokens for deep fusion. Unlike current MSA research, \ours positions fusion depth, scalability and dedicated multimodal capacity allocation as necessary factors for effective multimodal fusion. Our framework consists of a multimodal encoder that supplies a pretrained LM backbone with audiovisual information. A small set of fusion tokens is appended at the LM input and progressively: 1) gather linguistic information via LM blocks and 2) interact with information from the audiovisual encoder via MM Blocks. These MM Blocks are novel cross-modal cross-attention modules that are inserted after the LM layers (LM Blocks), enabling both deep and scalable fusion by design while accumulating multimodal information in the learnable tokens. \ours is coupled with a learning recipe which consists of modality-specific task losses, a language modelling loss, and an embedding regularization technique which acts as a language regularizer. These elements collectively form a multimodal deep fusion framework, that can be applied at any language-based multimodal scenario.

Our comprehensive experimental analysis reveals that deeper fusion schemes (5-7 layers) consistently outperform shallower approaches, challenging existing approaches in the field. Moreover, we demonstrate that a relatively small multimodal capacity (8-20 fusion tokens) achieves optimal performance, providing important insights for multimodal architecture design. Furthermore, we show that progressively learning multimodal representations consistently outperforms jointly learning all representations at once, highlighting the importance of fusion curriculum. 

We evaluate \ours across three MSA benchmarks covering different languages, dataset sizes, language distributions, and modality imbalance levels. Our recipe achieves state-of-the-art results across all datasets examined showcasing its applicability and versatility. Our ablation illustrates that the proposed MM Block design outperforms other alternatives and that removing any loss term component results in performance degradation, highlighting the synergetic nature of our learning recipe. Additionally, we find that the integrated language embedding augmentation consistently works better than existing approaches in the literature.

\subsection{Future Work}
In future work, we plan to extend \ours to additional tasks, domains, and modalities, including vision-language and audio-language models. A promising research direction is applying our approach, \ours, to self-supervised setups such as multimodal language modelling, where integration with other modalities and larger LLMs could establish a novel architectural paradigm for multimodal learning. Further research will also explore in-depth analysis and utilization of learnable tokens in the generative process, and integration with other generative frameworks, such as diffusion models, where these tokens can serve as multimodal latents.



%







\ifCLASSOPTIONcaptionsoff
  \newpage
\fi



\bibliographystyle{IEEEtran}
\bibliography{msagpt}
\end{document}